\documentclass{article}

\usepackage[final]{neurips_2022}

\usepackage[utf8]{inputenc} 
\usepackage[T1]{fontenc}    
\usepackage{hyperref}       
\usepackage{url}            
\usepackage{booktabs}       
\usepackage{amsfonts}       
\usepackage{nicefrac}       
\usepackage{microtype}      
\usepackage{xcolor}         
\usepackage{natbib}         
\usepackage{graphicx}       
\usepackage{multirow}       

\title{The Effect of Alignment Objectives on Code-Switching Translation \footnote{}}

\author{
  Mohamed Anwar \\
  African Institute for Mathematical Sciences (AIMS) \\
  \texttt{mghanem@aimsammi.org} \\
}

\begin{document}

\maketitle

\begin{abstract}
One of the things that need to change when it comes to machine translation is the models' ability to translate code-switching content, especially with the rise of social media and user-generated content. In this paper, we are proposing a way of training a single machine translation model that is able to translate monolingual sentences from one language to another, along with translating code-switched sentences to either language. This model can be considered a bilingual model in the human sense. For better use of parallel data, we generated synthetic code-switched (CSW) data along with an alignment loss on the encoder to align representations across languages. Using the WMT14 English-French (En-Fr) dataset, the trained model strongly outperforms bidirectional baselines on code-switched translation while maintaining quality for non-code-switched (monolingual) data.
\end{abstract}
\section{Introduction}
It's very difficult to say exactly how many people in the world speak two languages fluently. However, some estimates predict that around 60\% of people worldwide are at least bilingual. And according to \textcolor{blue}{\cite{pawlikova1990multilingualism}}, Africa is the most multilingual continent in the world, which requires rethinking what we expect from our machine translation models when they're deployed especially with the rise of social media and user-generated content worldwide. \textcolor{blue}{\cite{rijhwani2017estimating}} has collected around 50 million unique tweets and found out that English-Spanish, English-French, and English-Portuguese were the three most commonly mixed pairs accounting for 21.5\%, 20.8\% and 18.4\% of all collected code-switched tweets respectively.

One of the things that need to change when it comes to machine translation is the models' ability to translate code-switched content which remains understudied in natural language processing (NLP) in general and machine translation (MT) in particular. Code-switched (CSW) denotes the alternation of two languages within a single utterance (\textcolor{blue}{\cite{poplack1980sometimes}}. It is a common communicative phenomenon that occurs in multilingual communities during spoken and written interactions.

The Matrix Language Frame (MLF) theory (\textcolor{blue}{\cite{myers1997duelling}}) defines the concept of matrix and embedded languages where the matrix language is the main language that the sentence structure should conform to and notably provides the syntactic morphemes, while the influence of the embedded language is lesser and is mostly manifested in the insertion of content morphemes.

Neural machine translation (NMT) has significantly improved the quality of machine translation in recent years (\textcolor{blue}{\cite{sutskever2014sequence}}; \textcolor{blue}{\cite{bahdanau2014neural}}; \textcolor{blue}{\cite{zhang2015deep}}; \textcolor{blue}{\cite{wu2016google}}; \textcolor{blue}{\cite{gehring2017convolutional}}; \textcolor{blue}{\cite{vaswani2017attention}}). In this paper, we aim to build a bidirectional neural machine translation (NMT), which is a single model that is able to translate in both directions of the two languages it was trained on and is capable of translating code-switched sentences. In a sense, we can call this model a "Bilingual Translation Model" in the way humans are bilingual. When a human is bilingual, it means he/she is able to comprehend two languages and any combination of them. So, in a sense, this model is one step closer to having a true bilingual machine translation model (\textcolor{blue}{\cite{anwar2022true}}).

Training machine translation models on code-switched data is difficult due to a lack of data and a complete understanding of how humans code-switch. In this paper, we are proposing a method of generating code-switched sentences by leveraging parallel data. This method was highly inspired by the Masked Language Modeling (MLM) pre-training task in BERT (\textcolor{blue}{\cite{devlin2019bert}}) and adapted from the work of \textcolor{blue}{\cite{xu2021can}}.

Using only parallel data accompanied by synthetic code-switched data, the NMT model learns how to combine words from parallel sentences and identifies when to switch from one language to the other. Moreover, it captures code-switched constraints by attending and aligning the words in inputs, without requiring any external knowledge. To improve the performance even more, we propose an alignment objective applied only on the encoder side to incentivize the encoder to create language-agnostic representations making the decoder's job a little bit easier based on our experiments.

Our main contribution to this paper can be summarized into the following:

\begin{itemize}
    \item A new method of generating code-switched data using only parallel data (Section \textcolor{blue}{\ref{section:csw_data}}).
    \item An analysis of the generated code-switched data (Section \textcolor{blue}{\ref{section:analysis}}).
    \item An alignment objective implemented on the encoder-side to motivate the encoder to generate language-agnostic representations leveraging parallel \& code-switched data (Section \textcolor{blue}{\ref{section:alignment}}).
    \item A bilingual neural machine translation model that is almost as good as our bidirectional baseline model, while achieving better performance than our code-switched baseline model (Section \textcolor{blue}{\ref{section:exps}}).
\end{itemize}
\section{Related Work}

Research in the area of NLP on code-switching (CSW) has mostly focused on Language Modeling, especially for Automatic Speech Recognition (ASR) (\textcolor{blue}{\cite{pratapa2018language}}; \textcolor{blue}{\cite{garg2018code}}; \textcolor{blue}{\cite{gonen2018language}}; \textcolor{blue}{\cite{winata2019code}}; \textcolor{blue}{\cite{lee2020modeling}}). Evaluation tasks and benchmarks have also been prepared for LID in user-generated CSW content (\textcolor{blue}{\cite{zubiaga2016tweetlid}}; \textcolor{blue}{\cite{molina2019overview}}), Named Entity Recognition (\textcolor{blue}{\cite{aguilar2019named}}), Part-of-Speech tagging (\textcolor{blue}{\cite{ball2018part}}; \textcolor{blue}{\cite{aguilar2020lince}}; \textcolor{blue}{\cite{khanuja2020gluecos}}) and Sentiment Analysis (\textcolor{blue}{\cite{patwa2020semeval}}). CSW was also found useful in foreign language teaching: \textcolor{blue}{\cite{renduchintala2019simple}} showed that replacing words with their counterparts in a foreign language helps to learn foreign language vocabulary.

Regarding Machine Translation, most past work has focused on using synthetic code-switched (CSW) data to help conventional translation systems. \textcolor{blue}{\cite{huang2014improving}} used CSW corpus to improve word alignment and statistical MT. \textcolor{blue}{\cite{dinu2019training}} experienced replacing and concatenating source terminology constraints by the corresponding translation(s) to boost the accuracy of term translations. \textcolor{blue}{\cite{song2019code}} shared the same idea by replacing phrases with pre-specified translation to perform "soft" constraint decoding. A different line of research is in \textcolor{blue}{\cite{bulte2019neural}}, which explores ways to combine a source sentence with similar translations extracted from translation memories.

Similar to our work, \textcolor{blue}{\cite{yang2020csp}} proposed a new technique for NMT pre-training, which makes full use of the cross-lingual alignment information contained in source and target monolingual corpus. The idea is to randomly replace some words from the source language with their translations in the target language. This was done by using unsupervised word embedding mapping. Similar to that work, \textcolor{blue}{\cite{lin2020pre}} and \textcolor{blue}{\cite{pan2021contrastive}} create code-switching sentences by randomly replacing a word in the source language with another word in a different random language using MUSE (\textcolor{blue}{\cite{conneau2017word}}) as a pre-training step for big multilingual machine translation model. Also, \textcolor{blue}{\cite{gautam2021comet}} tried to translate from English to Hinglish (code-switched sentences between Hindi and English) by fine-tuning mBART (\textcolor{blue}{\cite{liu2020multilingual}}). Since mBART uses Hindi in Devanagari scripts, they converted Hindi from Roman scripts to Devanagari scripts. After getting a code-mixture of English-Hindu, they transliterated the Hindu from Devanagari to Roman scripts.

When it comes to generating code-switching data, \textcolor{blue}{\cite{gupta2020semi}} proposed a new way to generate synthetic code-switching data from parallel corpus by using a word-alignment such as fast-align (\textcolor{blue}{\cite{fast-align}}) algorithm to obtain the alignment matrix. Then, they used the Stanford library Stanza (\textcolor{blue}{\cite{qi2020stanza}}) to extract Part-of-Speech of the sentence (mainly adjective), Named entity (NE) of types: "Person", "location", and "organization", and Noun phrase (NP). After extracting, they inserted them into the appropriate places of the sentences in the other language.

Very similar to our way of generating code-switched data, \textcolor{blue}{\cite{xu2021can}} proposed a way of generating code-switched sentences using only fast-align (\textcolor{blue}{\cite{fast-align}}) algorithm to obtain the alignment matrix which can be used later for the generation. More details regarding this will be discussed later in this paper (Section \textcolor{blue}{\ref{section:csw_data}}).

\section{Data}
\label{section:data}
Since English and French are among the most high-resource spoken languages in the world and especially in Africa, we used the WMT14 English-French benchmark for training, newstest2008-2013 for validation, and newstest2014 for testing. To handle out-of-vocabulary words, we used a shared vocabulary of 40K Byte-Pair-Encoding (BPE) \textcolor{blue}{\cite{sennrich2016neural}} sub-words. 
We removed any sentences whose lengths are less than $1$ tokens and more than $250$ tokens.

\begin{figure}[ht]
  \centering
  \fbox{\includegraphics[scale=0.30, keepaspectratio]{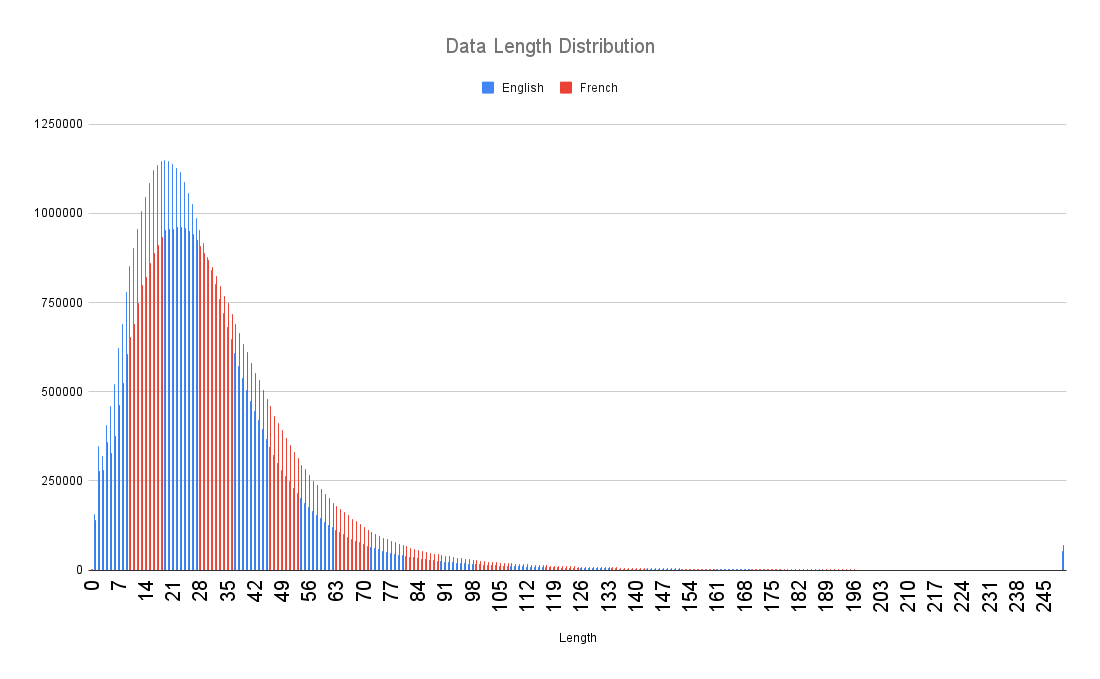}}
  \caption{Histogram of the sentence length of the WMT14 corpus.}
  \label{fig:wmt14_distribution}
\end{figure}

Also, we prepended a target-language tag to the sentences as shown in Table \textcolor{blue}{\ref{tab:data}} to teach the model to only care for the target language. All data were tokenized and normalized using Moses SMT toolkit (\textcolor{blue}{\cite{koehn2007moses}}). The WMT14 sentence length distribution before cleaning can be found in Figure \textcolor{blue}{\ref{fig:wmt14_distribution}}, and the stats of the data after cleaning can be found in Table \textcolor{blue}{\ref{tab:stats}}.

\begin{table}[h]
    \caption{WMT-14 French-English data stats used for training, validation, and test.}
    \centering
    \label{tab:stats}
    \begin{tabular}{l|lll|l}
    \toprule 
    Dataset & Train Size & Valid Size & Test Size & Total\\
    \midrule
          Fr-En & 35,789,717 & 15,827 & 3,003 & 35,808,547\\
          Bidirectional & 71,579,434 & 31,654 & 6,006 & 71,617,094\\
          CSW & 77,198,306 & 32,581 & 6,006 & 77,236,893\\
    \bottomrule
    \end{tabular}
\end{table}

\subsection{Code-switched (CSW) Data}
\label{section:csw_data}
Parallel corpora for code-switched data are very scarce (\textcolor{blue}{\cite{menacer2019machine}}). However, there have been previous works on generating synthetic code-switched data. Similar to \textcolor{blue}{\cite{song2019code}} and \textcolor{blue}{\cite{xu2021can}}, we created code-switched data by modifying the steps of \textcolor{blue}{\cite{xu2021can}}. Since \textcolor{blue}{\cite{xu2021can}} used only code-switched (CSW) data to train their models, they followed an exponential distribution \footnote{The number of replacements $r$ to appear in a derived code-switched sentence follows the distribution defined by this equation: $P(r=k) = \frac{1}{2^{k+1}}$ where $k \in \{1, ..., rep\}$ such that $rep$ is a predefined hyper-parameter that represents the maximum number of replacements.} to generate code-switched data that prefers smaller number of replacements when generating. Using this exponential distribution, they would generate code-switched sentences with 0 replacements around 50\% of the time, and with 1 replacement around 25\%, ...etc.

In our case, we decided to do things a little differently. We would rather have a separate group of code-switched sentences and mix it with another group of monolingual data when needed. To generate the code-switched data, we were looking for a distribution that increases the number of replacements with longer sentences, which is what humans tend to do. We can achieve that using linear distribution, but what is the reasonable slope for that?

We found the answer in the Masked Language Modeling (MLM) pre-training task used by BERT (\textcolor{blue}{\cite{devlin2019bert}}), where code-switching will be used instead of masking. In other words, we are going to code-switch the matrix language text with the embedded language text randomly around 15\% tokens of the whole input sentence. In that way, we can study the effect of code-switching and also can increase the number of replacements with longer sentences.

The following are the steps -in more detail- that we followed to generate the code-switched (CSW) data; a sample of the generated code-switching data can be found in Table \textcolor{blue}{\ref{csw_table}}.

\begin{enumerate}
    \item \textbf{Data Pre-processing}: First, we cleaned, tokenized, and normalized the text data using \emph{clean-corpus-n.perl} script from the moses SMT (\textcolor{blue}{\cite{koehn2007moses}}) toolkit.
    
    \item \textbf{Word Aligning}: Then, we started extracting word alignment using \emph{fast-align} toolkit (\textcolor{blue}{\cite{fast-align}}) with \emph{gdfa} (grow-diag-final-and).
    
    \item \textbf{Minimal Aligned Units}: Then, we extracted minimal alignment units following the approach of \textcolor{blue}{\cite{song2019code}} and \textcolor{blue}{\cite{xu2021can}}. These correspond to word segments $(a, b)$ extracted from the word alignments in the previous step; such that all alignment links outgoing from words in $a$ reach a word in $b$, and vice-versa.
    
    \item \textbf{Random Replacement}: Finally, aligned segments were replaced by under the following conditions:
    \begin{enumerate}
        \item Matrix (dominant) Language — defined by the Matrix Language Frame (MLF) theory (\textcolor{blue}{\cite{myers1997duelling}}) — is chosen randomly (50-50)\%.
        \item Similar to MLM pre-training (\textcolor{blue}{\cite{devlin2019bert}}), we randomly replaced around 15\% tokens of the sentence length in the matrix language with its aligned segments in the embedded language.
        \item Short sequences (less than 7 tokens) will have just one replacement.
        \item Positions of aligned segments were chosen uniformly.
    \end{enumerate}
\end{enumerate}
 
We combined the code-switched data generated with the parallel data after prepending a target language token (\textcolor{blue}{\cite{johnson2017googles}}). Table \textcolor{blue}{\ref{tab:data}} shows an example of the formulation of our data, where the \textbf{bold} words are in French and the normal words are in English. The first two sentences are monolingual while the last two examples are code-switching sentences. The first and last sentences are being translated to French, hence the \emph{<2fr>} language tag. The second and third sentences are being translated to English, hence the \emph{<2en>} language tag.
 
\begin{table}[ht]
  \caption{Example of how data looks like when mixing parallel data with generated code-switching data.}
  \label{tab:data}
  \centering
  \begin{tabular}{ll}
    \toprule
    Source & Target \\
    \midrule
    $<$2fr$>$ The weather today is nice. & \textbf{Il fait beau aujourd'hui .}\\
      $<$2en$>$ \textbf{Il fait beau aujourd'hui .} & The weather today is nice .\\
      $<$2en$>$ \textbf{Il fait} nice today . & The weather today is nice .\\
      $<$2fr$>$ \textbf{Il fait} nice today . & \textbf{Il fait beau aujourd'hui .}\\
    \bottomrule
  \end{tabular}
\end{table}

\begin{table}[ht]
    \caption{A sample of the code-switched data generated from the newstest2014 dataset. Bold words are French while normal ones are English.}
    \label{csw_table}
    \centering
    \begin{tabular}{p{3cm} p{3cm} p{3cm} p{1.5cm}}
    \toprule 
    CSW Sentence & English Translation & French Translation & Matrix Language \\
    \midrule\midrule
    Difficult Year \textbf{pour les} Pharmacists . & Difficult Year for Pharmacists . & Année difficile pour les pharmaciens . & English \\
    \midrule
    \textbf{Il ne} believe \textbf{pas que l’Ontario emboîtera le pas} . & He does not believe that Ontario will follow suit . & Il ne croit pas que l'Ontario emboîtera le pas . & French\\
    \midrule
    Asked how he had developed his character , the \textbf{acteur} and singer Justin Timberlake \textbf{avait rappelé} how he " \textbf{grandi dans} Tennessee , bathed in the blues and country music " . & When asked how he came up with his character, actor and singer Justin Timberlake recalled that he " grew up in Tennessee, surrounded by blues and country music ". & Interrogé sur la façon dont il a composé son personnage , l'acteur et chanteur Justin Timberlake avait rappelé avoir " grandi dans le Tennessee, baigné par le blues et la country " . & English\\
    \midrule
    \textbf{Mes camarades} cried with \textbf{joie et mes parents ont conservé} every {journaux qu’ils ont trouvés} . & My classmates cried with joy , and my parents saved every newspaper they could find . & Mes camarades de classe ont pleuré de joie, et mes parents ont gardé tous les journaux qu'ils ont pu trouver . & French\\
    \bottomrule
    \end{tabular}
\end{table}

\subsection{Analysis}
\label{section:analysis}

We used WMT-14 data for code-switching data generation and for training our machine translation systems. As a cleaning step, we removed all English and French sentences that were shorter than or equal to 1 token-long before using the proposed method (Section \textcolor{blue}{\ref{section:csw_data}}) to generate code-switching (CSW) data. After generation, we discarded sentence pairs with a source/target ratio higher than 1.5, with a maximum sentence length of 250. We performed the same procedure over the training, validation, and test sets generating around 77 million sentences, 32 thousand sentences, and 6 thousand sentences respectively. Exact numbers can be found in Table \textcolor{blue}{\ref{tab:stats}}.

\begin{figure}[ht]
  \centering
  \fbox{\includegraphics[scale=0.35, keepaspectratio]{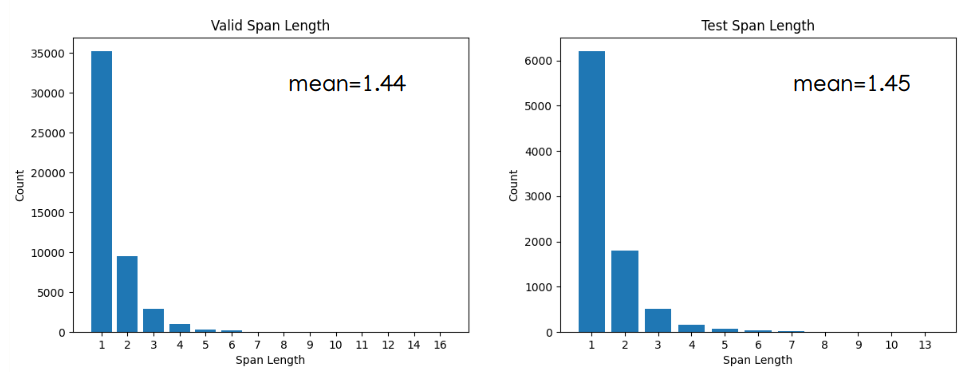}}
  \caption{Histogram of the span of the aligned units that were used for replacement when generating code-switching sentences of the validation set (left) and test set (right).}
  \label{fig:span}
\end{figure}

\begin{figure}[ht]
  \centering
  \fbox{\includegraphics[scale=0.35, keepaspectratio]{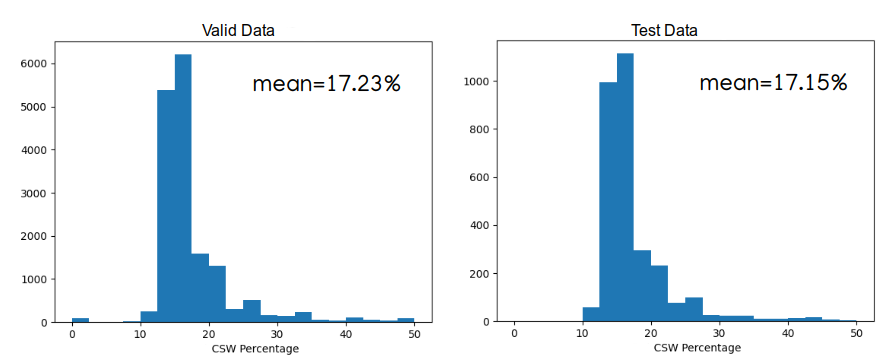}}
  \caption{Histogram of the generated code-switching sentences of the validation set (left) and test set (right).}
  \label{fig:replacement_distribution}
\end{figure}

The replacement distribution for the validation set and test set can be seen in Figure \textcolor{blue}{\ref{fig:replacement_distribution}}. As we can see, most of the replacements were around 15\% of the input monolingual sentences. Some of the generated sentences are between 15\%-20\% and that's because of the span of the minimal aligned units. As shown in Figure \textcolor{blue}{\ref{fig:span}}, the span of replaced units can be longer than one token, which counts for more percentage.

For example, let's assume that we have a sentence of 100 tokens, replacing 15\% means replacing around 15 tokens. But what if some of these replacements are more than one token long, this will count for more percentage. That's why some of the sentences were between 15\% and 20\%. Sentences below 15\% or above 20\% are usually short sentences. As seen from the data distribution shown in Figure \textcolor{blue}{\ref{fig:wmt14_distribution}}, a big portion of the sentences in both English and French languages are shorter than 10-tokens long. Replacing just two tokens in these accounts for more than 20\%, while replacing one counts for less than 10\%.

\section{Latent Alignment Objectives}
\label{section:alignment}
\textcolor{blue}{\cite{arivazhagan2019missing}} adapted an alignment objective to help with zero-shot translation by minimizing the discrepancy between the feature distributions of the source and target domains improving the machine translation model's performance on zero-shot translation. Inspired by that work, we decided to use the objective function with a slightly different goal in mind.

\begin{figure}[ht]
  \centering
  \fbox{\includegraphics[scale=0.35, keepaspectratio]{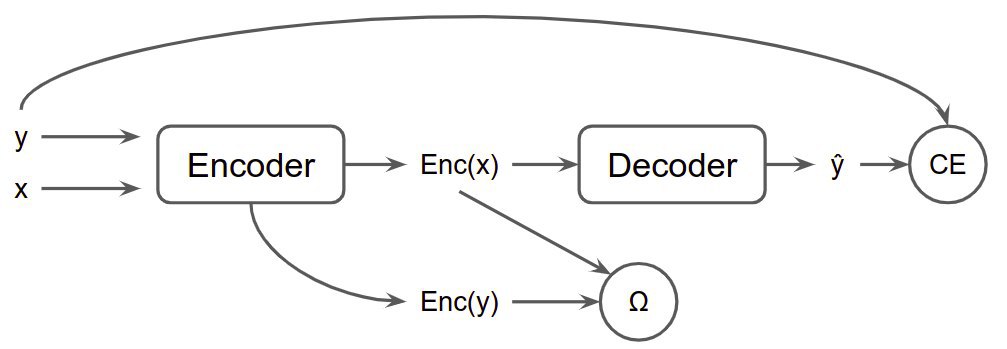}}
  \caption{shows the loss functions used with the encoder-decoder Transformer model, where $CE$ is the cross entropy, $\Omega$ is the alignment objective loss.}
  \label{fig:encoder_loss}
\end{figure}

As shown in Table \textcolor{blue}{\ref{tab:data}}, when mixing parallel data with code-switching data, we will have four sentence pairs that have the same meaning despite belonging to different distributions. To make the best use of that, we decided to use the alignment objective to improve bidirectional machine translation models' performance on code-switching translation by aligning the encoder's latent representation. In other words, we use the combination of parallel data and code-switching data and enforce the encoder to make language-agnostic representations about the input sentences that have the same meaning. Unlike \textcolor{blue}{\cite{arivazhagan2019missing}} where the whole model’s parameters were updated, we update only the encoder parameters, as shown in Figure \textcolor{blue}{\ref{fig:encoder_loss}}.

Below is an overview of the different objective functions that can be used to achieve that goal. Originally, these objectives were used to achieve different goals and fix different problems. However, they fit perfectly into our use case. Then in Section \textcolor{blue}{\ref{section:exps}}, we are going to discuss how we are going to use them to achieve our goal and explore their impact on the code-switching translation task, or at least some of them.

\subsection{Pool-Cosine Similarity}
\label{Pool-Cosine}

To improve the performance of multilingual NMT models on the zero-shot translation task, \textcolor{blue}{\cite{arivazhagan2019missing}} proposed viewing the zero-shot translation as a domain adaptation problem and use English - the language with which we always have parallel data - as the source domain, and the other languages collectively as the target domain. Then, train the model using formula \textcolor{blue}{\ref{eq:1}} to increase the cosine similarity between the latent representations of the English sentences and all other translations in the other languages, which eventually improves the model's performance on zero-shot translation.

\begin{equation} \label{eq:1}
    \Omega = - \mathop{\mathbb{E}}_{x_{src}, x_{tgt}\sim D_{(en,T)}}[sim(Enc(x_{src}), Enc(x_{tgt}))]
\end{equation}

Where $\Omega$ is the alignment objective function, $D_{(en,T)}$ is the training data containing parallel language pairs of English and other target languages, $x_{src}$ is the source sentence while $x_{tgt}$ is the target, $Enc(x)$ is the max-pooled encoder representation of a sentence $x$ similar to \textcolor{blue}{\cite{gouws2016bilbowa}} and \textcolor{blue}{\cite{coulmance2016transgram}}, and $sim$ is the cosine-similarity.

We can benefit from this objective function by combining our parallel data and mixing them with the generated code-switching data and using them as training data. In that case, we are going to train the encoder to maximize the cosine similarity between monolingual or code-switched sentences that have the same translations as monolingual ones found in the parallel data

\subsection{Cosine Similarity with Negative Sampling}

Very similar to Pool-Cosine Similarity (\textcolor{blue}{\ref{Pool-Cosine}}), \textcolor{blue}{\cite{wieting2019simple}} proposed an objective function where for each sentence pair $(x_i,y_i)$ during training, they randomly chose a negative target sentence $y_i'$ that is not a translation of $x_i$ and then train the model to ensure that source and target
sentences would be more similar than source and negative target examples by a margin $\delta$ as shown in the following equation:

\begin{equation} \label{eq:2}
    \Omega = \mathbb{E} [\delta - sim(Enc(x_{i}), Enc(y_{i})) + sim(Enc(x_{i}), Enc(y'_{i})) ]
\end{equation}

Unlike Pool-Cosine Similarity (\textcolor{blue}{\ref{Pool-Cosine}}), they used different encoder for each language, also the sentence representation $Enc()$ was achieved by averaging the embeddings of the subword units instead of max-pooling the last layer of the encoder representations.

\subsection{ Translation Ranking Objective }
\label{section:ranking}
Given a sentence-pair $(x,y)$ that are translations of each other, the translation candidate ranking task attempts to rank the true translation $y$ over all other target sentences $Y$. \textcolor{blue}{\cite{guo2018effective}} introduced a scoring function $\phi$ that assesses the compatibility between $x$ and $y$, and then trained the model to maximize the similarity between the correct sentence pairs normalized by summation of the similarity scores between the source sentence and the wrong $K$ target sentences, as shown in the following formula:

\begin{equation} \label{eq:3}
    \Omega = - \mathbb{E} \left[\frac{e^{\phi(x,y)}}{\sum_{k=1}^{K} e^{\phi(x,y)}} \right]
\end{equation}

Where they defined $\phi$ to be the dot product of sentence embeddings for the source $u$ and the target $v$, with $\phi(x,u) = u^\top \cdot v$.

When adapting this function to our use case, we can use this similarity metric to make monolingual and code-switched sentences more similar to their correct translations, normalized by other wrong translations.

\subsection{ AMS objective with Dual Encoder }

\textcolor{blue}{\cite{yang2019improving}} extend \textcolor{blue}{\cite{guo2018effective}} approach -explained in section \textcolor{blue}{\ref{section:ranking}}- by using a bidirectional dual-encoder with additive margin Softmax (\textcolor{blue}{\cite{wang2018additive}}) objective introducing a large margin, $m$, around positive pairs, which significantly improves the model performance as shown in the following figure:

\begin{equation} \label{eq:4}
    \Omega = - \mathbb{E} \left[\frac{e^{\phi(x,y) - m}}{ e^{\phi(x,y)-m} + \sum_{k=1}^{K} e^{\phi(x,y) }} \right]
\end{equation}

\subsection{ Sentence Alignment Objective }
\label{sa}
\textcolor{blue}{\cite{AMBER}} proposed an objective to encourage cross-lingual alignment of sentence representations in language modeling. For a source-target sentence pair $(x, y)$ in the parallel corpus, they separately calculated sentence embeddings denoted as $c_x$, $c_y$ by averaging the embeddings in the final layer as the sentence embeddings. Then, they encourage the model to predict the correct translation $y$ given a source sentence $x$. To do so, they define the sentence alignment loss as the average negative log-likelihood of the conditional probability of a candidate sentence $y$ being the correct translation of a source sentence $x$; as shown in the following formula:

\begin{equation} \label{eq:5}
    \Omega = - \mathbb{E} \left[log \frac{e^{c_x^\top \cdot c_y}}{ \sum_{y' \in \mathcal{M} \cup \mathcal{P}} e^{c_x^\top \cdot c_y'} } \right]
\end{equation}

Where $y'$ can be any sentence in any language that is sampled within a mini-batch of either monolingual data $\mathcal{M}$ or parallel data $\mathcal{P}$.

Same as earlier objectives, this can be suitable for our use-case when using our code-switching data instead of the monolingual data.

Since these different objective functions look very similar when considering our goal and since we have limited computation resources, we decided to experiment with only two of these objectives; namely the "\nameref{Pool-Cosine}" and the "\nameref{sa}". In the next section \textcolor{blue}{\ref{section:exps}}, we are going to do so and discuss their impact on code-switching translation.

\section{Experiments and Results}
\label{section:exps}
In all of our experiments, we used Transformer-Base (\textcolor{blue}{\cite{vaswani2017attention}}) configuration implemented in the Fairseq (\textcolor{blue}{\cite{ott2019fairseq}}) framework with 6 encoder layers and 6 decoder layers, each layer has a hidden size of $512$, an $8$ attention heads, where each attention head has a size of $64$, and a feed-forward hidden size of $2048$.

All models used in our experiments were trained using Adam optimizer (\textcolor{blue}{\cite{kingma2014adam}}) with $\beta_1=0.9, \beta_2=0.999$ and $\epsilon = 10^{-9}$. We started with a learning rate of $5e-4$, and warmed up the learning rate over the first $4000$ steps using inverse square root decay. For regularization, we used three different techniques:

\begin{itemize}
    \item Dropout (\textcolor{blue}{\cite{srivastava2014dropout}}) of $0.1$.
    \item Weight decay or L2 regularization (\textcolor{blue}{\cite{cortes2012l2}}) of $0.0001$.
    \item Label smoothing (\textcolor{blue}{\cite{szegedy2016rethinking}}) of $0.1$.
\end{itemize}

For training these models, we used four Tesla T400 GPUs. For handling out-of-vocabulary (OOV) tokens, we used a shared vocabulary of 40K BPE \textcolor{blue}{\cite{sennrich2016neural}} sub-words as mentioned earlier, and a batch size of $4096$ tokens per batch. The full list of the model's hyper-parameters can be found in Table \textcolor{blue}{\ref{tab:hyperparameters}}.
\begin{table}[h]
    \centering
    \caption{The hyperparameter values setting for training.}
    \label{tab:hyperparameters}
    \begin{tabular}{ll}
    \toprule 
    Hyper-parameter & Value\\
    \midrule
          Number of Layers & 6\\
          Hidden size & 512\\
          FFN inner hidden size & 2048\\
          Attention heads & 8\\
          Attention head size & 64\\
          Dropout & 0.1\\
          Attention Dropout & 0.0\\
          Warmup Steps & 4000\\
          Learning Rate & 5e-4\\
          Learning Rate Decay & inverse\_sqrt\\
          Batch Size & 4096 tokens\\
          Label Smoothing & 0.1\\
          Weight Decay & 0.0001\\
          Adam $\epsilon$ & $10^{-9}$\\
          Adam $\beta_1$ & 0.9\\
          Adam $\beta_2$ & 0.98\\
          Alignment Objective Weight & 1 (for SA) and 10 (for cosine)\\
    \bottomrule
    \end{tabular}
\end{table}

For our experimentation, we created two baselines using the same training hyper-parameters detailed in Table \textcolor{blue}{\ref{tab:hyperparameters}}:

\begin{enumerate}
    \item \textbf{Bidirectional}: A $En \leftrightarrow Fr$ bidirectional model trained only on parallel data.
    \item \textbf{csw}: A $En \leftrightarrow Fr$ bidirectional model trained only on synthetic code-switched data.
\end{enumerate}

 Then, we trained three other models:
 
 \begin{enumerate}
     \item \textbf{bi+csw}: A $En \leftrightarrow Fr$  bidirectional model that was trained on both parallel data and code-switched data without any alignment objectives.
     \item \textbf{bi+csw+cosine}: A $En \leftrightarrow Fr$  bidirectional model that was trained on both parallel data and code-switched data with the Pool-Cosine Similarity (\textcolor{blue}{\ref{Pool-Cosine}}) alignment objective.
     \item \textbf{bi+csw+sa}: A $En \leftrightarrow Fr$  bidirectional model that was trained on both parallel data and code-switched data with the Sentiment Alignment (\textcolor{blue}{\ref{sa}}) objective.
 \end{enumerate}
 
Table \textcolor{blue}{\ref{tab:results}} shows the results of all models trained using the same parameters seen in Table \textcolor{blue}{\ref{tab:hyperparameters}} where all models were trained till convergence with patience = 10 and the reported results are case-sensitive detokenized 4-gram BLEU (\textcolor{blue}{\cite{papineni2002bleu}}) Scores with a beam size = 5.

\begin{table}[ht]
  \caption{Case-sensitive detokenized 4-gram BLEU Score on unidirectional and CSW data from newstest2014 using SacreBLEU (\textcolor{blue}{\cite{post2018bleuclarity}}) with beam-size of 5.}
  \label{tab:results}
  \centering
  \begin{tabular}{llllll}
    \toprule
    \multirow{2}{*}{Model} & \multirow{2}{*}{Steps} & \multicolumn{2}{c}{Unidirectional} & \multicolumn{2}{c}{Code-switching} \\
    \cmidrule(r){3-4} \cmidrule(r){5-6}
    & & En $ \rightarrow $ Fr & Fr $ \rightarrow $ En & CSW $ \rightarrow $ Fr & CSW $ \rightarrow $ En \\
    \midrule
    Bidirectional & 642K & \bf{39.57} & \bf{36.17} & 57.86 & 60.77 \\
    csw & 594K & 8.38 & 13.66 & 68.49 & 66.65 \\
    bi+csw & 420k & 38.57 & 34.75 & 68.38 & 66.31 \\
    bi+csw+cosine & 612k & 39.18 & 35.43 & \bf{68.69} & \bf{66.96} \\
    bi+csw+sa & 660k & 39.00 & 35.68 & 68.58 & 66.88 \\
    \bottomrule
  \end{tabular}
\end{table}

From Table \textcolor{blue}{\ref{tab:results}}, we see that the \textbf{Bidirectional} baseline performs well for bi-directional translation. However, \textbf{csw} baseline performs well only on CSW translation.

Our trained models \textbf{bi+csw}, \textbf{bi+csw+cosine} and \textbf{bi+csw+sa} work well across the board where \textbf{bi+csw+cosine} has the best performance on CSW data while achieving competitive results on bi-directional translation compared to other baselines.

To further understand the effect of these alignment objectives on the code-switching translation, we've decided to split the code-switched test data into four groups based on the target language and whether it's a matrix or embedded language. So, these four groups can be seen in Table \textcolor{blue}{\ref{tab:mat_embd}}, and they are:

\begin{enumerate}
    \item \textbf{CSW $\rightarrow$ Fr (Matrix)}: Target is French and Matrix language is French.
    \item \textbf{CSW $\rightarrow$ Fr (Embedded)}: Target is French and Embedded language is French.
    \item \textbf{CSW $\rightarrow$ En (Matrix)}: Target is English and Matrix language is English.
    \item \textbf{CSW $\rightarrow$ En (Embedded)}: Target is English and Embedded language is English.
\end{enumerate}

\begin{table}[ht]
  \caption{Case-sensitive detokenized 4-gram BLEU Score on code-switching data, generated from newstest2014 and scored with  (\textcolor{blue}{\cite{post2018bleuclarity}}) with beam-size of 5.}
  \label{tab:mat_embd}
  \centering
  \begin{tabular}{llllll}
    \toprule
    \multirow{2}{*}{Model} & \multirow{2}{*}{Steps} & \multicolumn{2}{c}{CSW $ \rightarrow $ Fr} & \multicolumn{2}{c}{CSW $ \rightarrow $ En} \\
    \cmidrule(r){3-4} \cmidrule(r){5-6}
    & & Matrix & Embedded & Matrix & Embedded \\
    \midrule
    copying & - & 65.34 & 7.91 & 66.80 & 7.66 \\
    Bidirectional & 642K & 71.67 & 42.39 & 81.04 & 40.43 \\
    csw & 594K & \bf{88.99} & 45.26 & \bf{89.03} & 43.82 \\
    bi+csw & 420k & 88.45 & 45.65 & 88.63 & 43.52 \\
    bi+csw+cosine & 612k & 88.85 & \bf{45.98} & 88.93 & \bf{44.63} \\
    bi+csw+sa & 660k & 88.84 & 45.62 & 89.12 & 44.20 \\
    \bottomrule
  \end{tabular}
\end{table}

Table \textcolor{blue}{\ref{tab:mat_embd}} shows that models trained with the alignment objectives scores very similar scores to the \textbf{csw} baseline on the matrix-language side, while achieving better scores on the embedded-language side. Also, we can see that these models do more than just copying since the scores for \textbf{copying}, which copies the code-switched source sentence and use it as the hypothesis, are incredibly bad especially on the embedded-language side.

\section{Conclusion \& Future Work}
In this paper, we introduced two ways to make best use of parallel data that can improve the model’s performance on both unidirectional and code-switched data. First, a statistical way to generate code-switched data that can be aggregated with parallel data for training. Second, a loss function that trains the encoder to generate language-independent representations. We show that these two techniques boosted our model’s performance on both unidirectional and code-switched translation.

This is still a work in progress, and we are exploring new ways to improve our model even more; and more importantly experimenting with another language family like Arabic, one of the most spoken languages in Africa.
\begin{ack}

I would like to express my deepest appreciation to the "Google Cloud credits for Academic Research" grant which provided the computational resources needed to perform our experimentation needed for this paper.

Also, I want to extend my appreciation to Jitao Xu and Yvon Francis for sharing their code, which we tried to improve upon to generate our synthetic CSW data.

Finally, I’m deeply indebted to my two supervisors: \textbf{Julia Kreutzer} (\texttt{jkreutzer@google.com}) and \textbf{Melvin Johnson} (\texttt{melvinp@google.com}) from Google Research who kept me sharp, motivated, and hungry for more.

\end{ack}

\bibliography{08-ref}
\bibliographystyle{plainnat}

\end{document}